\documentclass[11pt,a4paper]{article}

\usepackage[utf8]{inputenc}
\usepackage[T2A,T1]{fontenc}
\usepackage[ukrainian,english]{babel}
\usepackage{amsmath,amssymb}
\usepackage{graphicx}
\usepackage{booktabs}
\usepackage{multirow}
\usepackage{hyperref}
\usepackage{xcolor}
\usepackage{natbib}
\usepackage{url}
\usepackage{microtype}
\usepackage{enumitem}
\usepackage{caption}
\usepackage[margin=1in]{geometry}
\usepackage{float}
\usepackage{array}
\usepackage{pgfplots}
\usepackage{pgfplotstable}
\pgfplotsset{compat=1.18}

\hypersetup{
    colorlinks=true,
    linkcolor=blue!70!black,
    citecolor=blue!70!black,
    urlcolor=blue!70!black,
}

\newcolumntype{R}[1]{>{\raggedleft\arraybackslash}p{#1}}

\title{Few-Shot Degradation Is Not What It Seems:\\
Behavioral Evidence, Representation Analysis,\\
and a Random-Text Control Across 12 Models, 2 Tasks, and 2 Architectures}

\author{
  Volodymyr Ovcharov\thanks{Corresponding author: \texttt{volodymyr@legal.org.ua}} \\
  LEX AI Platform, legal.org.ua \\
  Kyiv, Ukraine
}

\date{May 2026}

\begin{document}

\maketitle

\begin{abstract}
When few-shot prompting degrades a language model, the intuitive explanation is \emph{distortion}: demonstrations shift the model's internal representations away from the correct answer, and more shift means more harm. We show this explanation is wrong. Evaluating 12~open-weight models on two tasks (news classification and Ukrainian legal case outcome prediction), we find that raw representation shift does not predict whether few-shot helps or hurts ($r = 0.20$). The reason is a confound that prior work overlooks: few-shot prompts are 5--8$\times$ longer than zero-shot prompts, and that length difference alone accounts for 40--79\% of the observed shift. We introduce a \emph{random-text control} -- replacing demonstrations with length-matched random tokens -- to isolate the shift caused by demonstration \emph{content} from the shift caused by prompt \emph{length}. The residual, which we call \emph{content delta}, reverses the picture: models whose representations change more because of what demonstrations say benefit more from few-shot prompting ($\rho = +0.65$, $p = 0.043$), not less. The model that degrades most (Llama~3.3~70B) has the lowest content delta among all Transformers -- its shift is almost entirely a length artifact. Zeroing its attention to demonstrations causally recovers accuracy above the zero-shot baseline. These findings establish the random-text control as a necessary methodological step for any study of in-context learning through representation geometry. Data: \url{https://huggingface.co/datasets/overthelex/attention-analysis-fewshot}.
\end{abstract}



\section{Introduction}

Why does few-shot prompting sometimes make language models worse? The natural explanation -- which we call the \emph{distortion hypothesis} -- is that demonstrations shift a model's internal representations away from the correct answer: more shift, more harm. This account is intuitive, consistent with the observation that degrading models show large representation changes under few-shot prompting, and implicit in work that treats representation shift as a diagnostic of interference \citep{hendel2023icl, wang2023label}.

We show it is wrong.

The distortion hypothesis fails because it ignores a confound: few-shot prompts are 5--8$\times$ longer than zero-shot prompts. This length difference alone moves representations substantially -- by an amount that varies across models from 14\% to 79\% of the total observed shift. Llama~3.3~70B, the model with the most severe few-shot degradation in our evaluation, has the highest proportion of length-driven shift (79\%) and the lowest proportion of content-driven shift among all Transformers. Its representations change a lot under few-shot prompting, but almost none of that change comes from what the demonstrations actually say.

To disentangle length from content, we introduce a \emph{random-text control}: we replace demonstrations with length-matched random tokens and measure the resulting shift. Subtracting this from the few-shot shift yields \emph{content delta} -- the representation change attributable to demonstration content alone. Content delta reverses the distortion prediction: models whose representations change more from demonstration content benefit more ($\rho = +0.65$, $p = 0.043$), not less. Zeroing Llama~3.3~70B's attention to demonstration tokens causally recovers accuracy above the zero-shot baseline, confirming that the demonstrations actively harm this model.

These findings have a methodological consequence: \textbf{any study that measures representation shift under few-shot prompting without controlling for prompt length risks drawing the opposite conclusion.} The random-text control is simple to implement (one additional forward pass per example) and should become standard practice.

We evaluate 12~open-weight models from 9~families on two tasks: news topic classification (SIB-200) and Ukrainian legal case outcome prediction \citep{ovcharov2026tokenizer}. \textbf{Part~I} (Section~\ref{sec:behavioral}) establishes behavioral scope: the few-shot effect is strongly task-dependent ($+$24~pp mean on news vs.\ $+$3.4~pp on legal), with two models degrading on at least one task. \textbf{Part~II} (Section~\ref{sec:distortion}) opens the models: we extract hidden states under zero-shot, few-shot, and random-text conditions, and show that the distortion account collapses once prompt length is controlled for.


\section{Related Work}

\subsection{Few-Shot Learning and the Distortion Assumption}

\citet{brown2020language} established few-shot in-context learning as a core LLM capability. Subsequent work revealed its fragility: performance depends on example selection \citep{liu2022makes}, format consistency \citep{min2022rethinking}, label distribution \citep{zhao2021calibrate}, and model scale \citep{wei2023larger}. \citet{ovcharov2026tokenizer} documented systematic few-shot degradation on Ukrainian legal text.

When few-shot prompting fails, the implicit explanation in the literature is representational distortion. \citet{hendel2023icl} showed that ICL compresses demonstrations into task vectors in hidden states; if this compression is noisy, the task vector could mislead rather than help. \citet{wang2023label} demonstrated that label words serve as attention anchors, creating a pathway for demonstration content to distort the classification decision. Neither work explicitly tests whether the magnitude of representation shift predicts degradation -- an assumption we refute.

\subsection{Representation Geometry of In-Context Learning}

A growing body of work analyzes ICL through the lens of representation geometry. \citet{dai2023gpt} formalized ICL as implicit gradient descent, predicting that demonstrations should move representations toward a task-specific optimum. \citet{todd2024function} discovered function vectors -- directions in activation space that encode task mappings and can be extracted and transplanted between prompts. \citet{olsson2022context} identified induction heads as the circuit-level mechanism for ICL.

These accounts predict that representation change under ICL is \emph{functional}: it encodes task-relevant information. The distortion hypothesis predicts the opposite -- that change is harmful. Our random-text control resolves this tension by showing that most of the observed ``distortion'' is actually a prompt-length artifact, and the content-driven residual is indeed functional.

\subsection{Attention and Architecture}

\citet{jain2019attention} cautioned that attention weights are not explanations; we confirm this empirically (DAR shows zero correlation with few-shot benefit). For SSMs, \citet{park2024mamba} showed that Mamba matches Transformers on regression ICL tasks but falls short on retrieval, motivating our architecture comparison. \citet{liu2024quantized} found that 4-bit quantization preserves emergent abilities including ICL, though degradation patterns vary by task -- relevant to our quantized Llama~3.3~70B results.


\section{Experimental Setup}
\label{sec:setup}

\subsection{Models}

We evaluate 12~open-weight models from 9~architecture families (Table~\ref{tab:models}): 11~Transformers and 1~SSM (Falcon Mamba~7B). Models range from 7B to 70B parameters; Llama~3.3~70B uses 4-bit NF4 quantization. All use native chat templates with temperature~0 (greedy decoding). For Qwen3 models (8B, 14B, 32B), thinking mode was disabled (\texttt{enable\_thinking=false}). Inference used \texttt{transformers}~4.51 and \texttt{torch}~2.6 on 4$\times$A10G GPUs (96~GB total). Model identifiers are listed in Appendix~\ref{app:models}.

\begin{table}[t]
\centering
\small
\caption{Models evaluated. Trans = Transformer, SSM = state-space model.}
\label{tab:models}
\begin{tabular}{lllr}
\toprule
\textbf{Model} & \textbf{Family} & \textbf{Arch} & \textbf{Size} \\
\midrule
Qwen3 32B / 14B / 8B   & Qwen      & Trans & 8--32B \\
Llama 3.3 70B           & Meta      & Trans & 70B \\
Mistral Nemo 12B / 7B   & Mistral   & Trans & 7--12B \\
Nemotron 8B             & NVIDIA    & Trans & 8B \\
Falcon3 7B              & TII       & Trans & 7B \\
Granite 3.1 8B          & IBM       & Trans & 8B \\
DeepSeek R1 14B         & DeepSeek  & Trans & 14B \\
Gemma 3 12B             & Google    & Trans & 12B \\
Falcon Mamba 7B         & TII       & \textbf{SSM} & 7B \\
\bottomrule
\end{tabular}
\end{table}

\subsection{Datasets}

\paragraph{SIB-200} \citep{adelani2024sib200}: 204 parallel test examples in 7~categories (news topic classification). We use Ukrainian (\texttt{ukr\_Cyrl}).

\paragraph{ua-case-outcome} \citep{ovcharov2026tokenizer}: Ukrainian court decisions with 7-class outcome labels. We use 196~stratified test examples (28~per class) with facts truncated to 1,500~characters and few-shot demonstrations truncated to 300~characters to fit context windows of smaller models.

\subsection{Internal Metrics}
\label{sec:metrics}

We define three metrics for analyzing model internals under different prompt conditions.

\paragraph{Representation shift.} Cosine distance between hidden states at the classification position in zero-shot vs.\ few-shot mode:
\begin{equation}
\text{shift}_{\text{FS}} = 1 - \cos(\mathbf{h}_{\text{ZS}}, \mathbf{h}_{\text{FS}})
\label{eq:shift}
\end{equation}
This is the quantity that the distortion hypothesis predicts should correlate negatively with few-shot benefit.

\paragraph{Random-text control.} We replace demonstrations with length-matched random text (preserving prompt structure and length) and measure the resulting shift:
\begin{equation}
\text{shift}_{\text{rand}} = 1 - \cos(\mathbf{h}_{\text{ZS}}, \mathbf{h}_{\text{rand}})
\label{eq:shift_rand}
\end{equation}
This captures the representation change caused by prompt length alone.

\paragraph{Content delta.} The difference isolates the effect of demonstration content:
\begin{equation}
\Delta_{\text{cnt}} = \text{shift}_{\text{FS}} - \text{shift}_{\text{rand}}
\label{eq:content_delta}
\end{equation}
If the distortion hypothesis is correct, $\Delta_{\text{cnt}}$ should correlate negatively with few-shot benefit. If the content hypothesis is correct, it should correlate positively.

\paragraph{Demonstration Attention Ratio (DAR).} Fraction of final-token attention (last 4 layers, averaged across heads) allocated to demonstration tokens vs.\ input tokens. Defined for Transformers only.


\section{Part I: Behavioral Results}
\label{sec:behavioral}

Table~\ref{tab:behavioral} presents accuracy on both tasks for all 12~models. The few-shot effect is strongly task-dependent: mean $\Delta_{\text{FS}}$ is $+$23.9~pp on SIB-200 news but only $+$3.4~pp on legal text -- a 7$\times$ reduction. On news, only 1/10~valid models degrades; on legal, 2/10 degrade.

\begin{table*}[t]
\centering
\small
\caption{Few-shot accuracy on SIB-200 (news, 204 examples) and ua-case-outcome (legal, 196 examples). Models sorted by legal $\Delta_{\text{FS}}$. Bold = degradation. ``fmt'' = invalid outputs.}
\label{tab:behavioral}
\begin{tabular}{llR{0.7cm}R{0.7cm}R{0.9cm}R{0.7cm}R{0.7cm}R{0.9cm}}
\toprule
& & \multicolumn{3}{c}{\textbf{SIB-200 (news)}} & \multicolumn{3}{c}{\textbf{ua-case-outcome (legal)}} \\
\cmidrule(lr){3-5} \cmidrule(lr){6-8}
\textbf{Model} & \textbf{Size} & \textbf{ZS} & \textbf{FS} & \textbf{$\Delta$} & \textbf{ZS} & \textbf{FS} & \textbf{$\Delta$} \\
\midrule
Falcon3 7B     & 7B  & 28.9 & 58.3 & $+$29.4$^{***}$ & 20.4 & 30.1 & $+$9.7$^{**}$ \\
Qwen3 8B       & 8B  & 35.8 & 73.0 & $+$37.3$^{***}$ & 24.0 & 33.7 & $+$9.7$^{**}$ \\
Mistral 7B     & 7B  &  6.9 & 54.4 & $+$47.5$^{***}$ & 29.6 & 34.7 & $+$5.1 \\
Gemma 3 12B    & 12B & 56.9 & 59.3 & $+$2.5 & 36.7 & 41.3 & $+$4.6 \\
Granite 8B     & 8B  & 33.3 & 56.9 & $+$23.5$^{***}$ & 30.1 & 33.2 & $+$3.1 \\
Qwen3 32B      & 32B & 57.8 & 72.5 & $+$14.7$^{***}$ & 38.8 & 40.3 & $+$1.5 \\
Nemotron 8B    & 8B  & 51.0 & 77.0 & $+$26.0$^{***}$ & 32.1 & 33.2 & $+$1.0 \\
Nemo 12B       & 12B & 32.4 & 58.3 & $+$26.0$^{***}$ & 33.2 & 34.2 & $+$1.0 \\
\textbf{Qwen3 14B} & 14B & 40.0 & 75.0 & $+$35.0$^{***}$ & 37.8 & 37.2 & $\mathbf{-0.5}$ \\
\textbf{Llama 3.3 70B} & 70B & 58.3 & 55.9 & $\mathbf{-2.5}$ & 35.7 & 34.7 & $\mathbf{-1.0}$ \\
DeepSeek R1 14B & 14B & \multicolumn{2}{c}{\footnotesize fmt} & --- & \multicolumn{2}{c}{\footnotesize fmt} & --- \\
Falcon Mamba 7B & 7B & \multicolumn{2}{c}{\footnotesize fmt} & --- & \multicolumn{3}{c}{} \\
\midrule
\multicolumn{2}{l}{\textit{Mean $\Delta$ (valid)}} & & & \textit{+24} & & & \textit{+3.4} \\
\multicolumn{2}{l}{\textit{Degrading / valid}} & & & \textit{1/10} & & & \textit{2/10} \\
\bottomrule
\multicolumn{8}{l}{\footnotesize McNemar's test: $^{*}p < 0.05$, $^{**}p < 0.01$, $^{***}p < 0.001$. No marker = tested, $p > 0.05$.}
\end{tabular}
\end{table*}

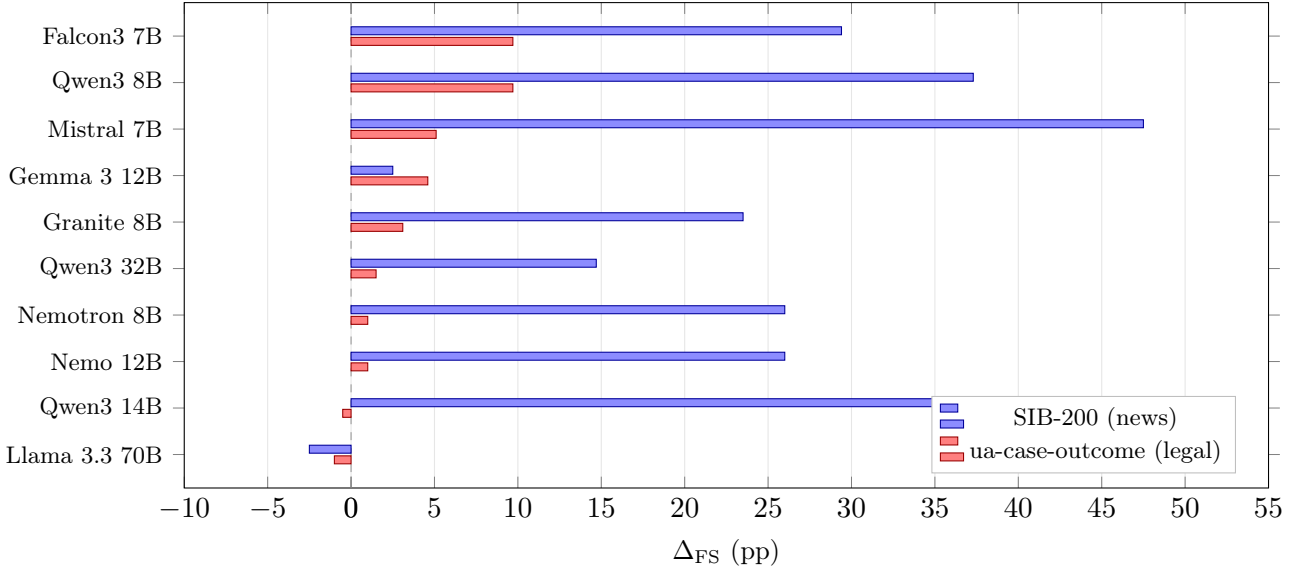
\begin{figure}[t]
\centering
\begin{tikzpicture}
\begin{axis}[
    width=\columnwidth,
    height=8cm,
    xbar,
    bar width=3pt,
    xlabel={$\Delta_{\text{FS}}$ (pp)},
    symbolic y coords={
        {Llama 3.3 70B},
        {Qwen3 14B},
        {Nemo 12B},
        {Nemotron 8B},
        {Qwen3 32B},
        {Granite 8B},
        {Gemma 3 12B},
        {Mistral 7B},
        {Qwen3 8B},
        {Falcon3 7B}
    },
    ytick=data,
    yticklabel style={font=\footnotesize},
    xlabel style={font=\small},
    xmin=-10, xmax=55,
    legend style={
        at={(0.97,0.03)},
        anchor=south east,
        font=\footnotesize,
        draw=gray!50,
    },
    enlarge y limits=0.08,
    xmajorgrids=true,
    ymajorgrids=false,
    grid style={gray!20},
    extra x ticks={0},
    extra x tick style={
        grid=major,
        grid style={black!40, thin, dashed},
    },
]
\addplot[fill=blue!45!white, draw=blue!60!black, bar shift=2pt] coordinates {
    (-2.5,{Llama 3.3 70B})
    (35.0,{Qwen3 14B})
    (26.0,{Nemo 12B})
    (26.0,{Nemotron 8B})
    (14.7,{Qwen3 32B})
    (23.5,{Granite 8B})
    (2.5,{Gemma 3 12B})
    (47.5,{Mistral 7B})
    (37.3,{Qwen3 8B})
    (29.4,{Falcon3 7B})
};
\addplot[fill=red!50!white, draw=red!60!black, bar shift=-2pt] coordinates {
    (-1.0,{Llama 3.3 70B})
    (-0.5,{Qwen3 14B})
    (1.0,{Nemo 12B})
    (1.0,{Nemotron 8B})
    (1.5,{Qwen3 32B})
    (3.1,{Granite 8B})
    (4.6,{Gemma 3 12B})
    (5.1,{Mistral 7B})
    (9.7,{Qwen3 8B})
    (9.7,{Falcon3 7B})
};
\legend{SIB-200 (news), ua-case-outcome (legal)}
\end{axis}
\end{tikzpicture}
\caption{Few-shot effect by task and model. Every model that benefits on news benefits less on legal text; two models degrade on legal (below dashed line). Models sorted by legal $\Delta_{\text{FS}}$.}
\label{fig:task_contrast}
\end{figure}

Several patterns emerge. First, the few-shot effect is strongly task-dependent: every model that improves on news improves less on legal. Second, Llama~3.3~70B degrades on \emph{both} tasks -- the only model with consistent cross-task degradation. These behavioral results motivate the central question: \emph{why} do demonstrations help some models but hurt others? The distortion hypothesis offers an intuitive answer: demonstrations corrupt internal representations. We test this directly in Part~II.


\section{Part II: Testing the Distortion Hypothesis}
\label{sec:distortion}

\subsection{The Prompt-Length Confound}
\label{sec:confound}

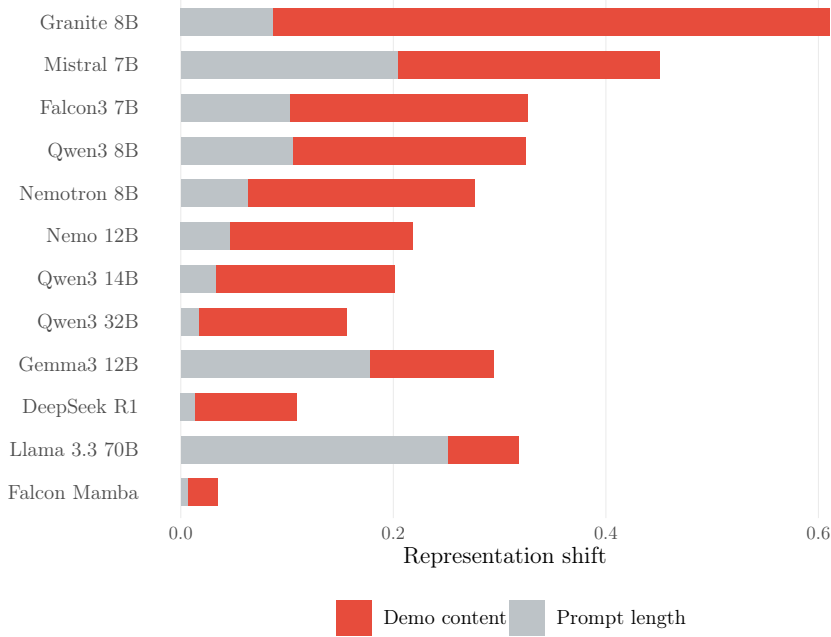
\begin{figure}[t]
\centering
\begin{tikzpicture}[x=1pt,y=1pt]
\definecolor{fillColor}{RGB}{255,255,255}
\path[use as bounding box,fill=fillColor,fill opacity=0.00] (0,0) rectangle (325.21,252.94);
\begin{scope}
\path[clip] (  0.00,  0.00) rectangle (325.21,252.94);
\definecolor{fillColor}{RGB}{255,255,255}

\path[fill=fillColor] (  0.00,  0.00) rectangle (325.21,252.94);
\end{scope}
\begin{scope}
\path[clip] ( 52.80, 52.77) rectangle (321.21,248.94);
\definecolor{drawColor}{gray}{0.92}

\path[draw=drawColor,line width= 0.4pt,line join=round] ( 65.00, 52.77) --
	( 65.00,248.94);

\path[draw=drawColor,line width= 0.4pt,line join=round] (144.98, 52.77) --
	(144.98,248.94);

\path[draw=drawColor,line width= 0.4pt,line join=round] (224.96, 52.77) --
	(224.96,248.94);

\path[draw=drawColor,line width= 0.4pt,line join=round] (304.94, 52.77) --
	(304.94,248.94);
\definecolor{fillColor}{RGB}{189,195,199}

\path[fill=fillColor] ( 65.00,234.07) rectangle ( 99.71,244.52);
\definecolor{fillColor}{RGB}{231,76,60}

\path[fill=fillColor] ( 99.71,234.07) rectangle (309.01,244.52);
\definecolor{fillColor}{RGB}{189,195,199}

\path[fill=fillColor] ( 65.00,217.99) rectangle (146.86,228.44);
\definecolor{fillColor}{RGB}{231,76,60}

\path[fill=fillColor] (146.86,217.99) rectangle (245.31,228.44);
\definecolor{fillColor}{RGB}{189,195,199}

\path[fill=fillColor] ( 65.00,201.91) rectangle (106.31,212.36);
\definecolor{fillColor}{RGB}{231,76,60}

\path[fill=fillColor] (106.31,201.91) rectangle (195.80,212.36);
\definecolor{fillColor}{RGB}{189,195,199}

\path[fill=fillColor] ( 65.00,185.83) rectangle (107.35,196.28);
\definecolor{fillColor}{RGB}{231,76,60}

\path[fill=fillColor] (107.35,185.83) rectangle (194.76,196.28);
\definecolor{fillColor}{RGB}{189,195,199}

\path[fill=fillColor] ( 65.00,169.75) rectangle ( 90.43,180.20);
\definecolor{fillColor}{RGB}{231,76,60}

\path[fill=fillColor] ( 90.43,169.75) rectangle (175.65,180.20);
\definecolor{fillColor}{RGB}{189,195,199}

\path[fill=fillColor] ( 65.00,153.67) rectangle ( 83.59,164.12);
\definecolor{fillColor}{RGB}{231,76,60}

\path[fill=fillColor] ( 83.59,153.67) rectangle (152.42,164.12);
\definecolor{fillColor}{RGB}{189,195,199}

\path[fill=fillColor] ( 65.00,137.59) rectangle ( 78.23,148.04);
\definecolor{fillColor}{RGB}{231,76,60}

\path[fill=fillColor] ( 78.23,137.59) rectangle (145.38,148.04);
\definecolor{fillColor}{RGB}{189,195,199}

\path[fill=fillColor] ( 65.00,121.51) rectangle ( 71.96,131.96);
\definecolor{fillColor}{RGB}{231,76,60}

\path[fill=fillColor] ( 71.96,121.51) rectangle (127.58,131.96);
\definecolor{fillColor}{RGB}{189,195,199}

\path[fill=fillColor] ( 65.00,105.43) rectangle (136.46,115.88);
\definecolor{fillColor}{RGB}{231,76,60}

\path[fill=fillColor] (136.46,105.43) rectangle (182.77,115.88);
\definecolor{fillColor}{RGB}{189,195,199}

\path[fill=fillColor] ( 65.00, 89.35) rectangle ( 70.40, 99.80);
\definecolor{fillColor}{RGB}{231,76,60}

\path[fill=fillColor] ( 70.40, 89.35) rectangle (108.59, 99.80);
\definecolor{fillColor}{RGB}{189,195,199}

\path[fill=fillColor] ( 65.00, 73.27) rectangle (165.69, 83.72);
\definecolor{fillColor}{RGB}{231,76,60}

\path[fill=fillColor] (165.69, 73.27) rectangle (192.36, 83.72);
\definecolor{fillColor}{RGB}{189,195,199}

\path[fill=fillColor] ( 65.00, 57.19) rectangle ( 67.96, 67.64);
\definecolor{fillColor}{RGB}{231,76,60}

\path[fill=fillColor] ( 67.96, 57.19) rectangle ( 78.91, 67.64);
\end{scope}
\begin{scope}
\path[clip] (  0.00,  0.00) rectangle (325.21,252.94);
\definecolor{drawColor}{gray}{0.30}

\node[text=drawColor,anchor=base east,inner sep=0pt, outer sep=0pt, scale=  0.70] at ( 49.20, 60.01) {Falcon Mamba};

\node[text=drawColor,anchor=base east,inner sep=0pt, outer sep=0pt, scale=  0.70] at ( 49.20, 76.09) {Llama 3.3 70B};

\node[text=drawColor,anchor=base east,inner sep=0pt, outer sep=0pt, scale=  0.70] at ( 49.20, 92.17) {DeepSeek R1};

\node[text=drawColor,anchor=base east,inner sep=0pt, outer sep=0pt, scale=  0.70] at ( 49.20,108.25) {Gemma3 12B};

\node[text=drawColor,anchor=base east,inner sep=0pt, outer sep=0pt, scale=  0.70] at ( 49.20,124.33) {Qwen3 32B};

\node[text=drawColor,anchor=base east,inner sep=0pt, outer sep=0pt, scale=  0.70] at ( 49.20,140.41) {Qwen3 14B};

\node[text=drawColor,anchor=base east,inner sep=0pt, outer sep=0pt, scale=  0.70] at ( 49.20,156.49) {Nemo 12B};

\node[text=drawColor,anchor=base east,inner sep=0pt, outer sep=0pt, scale=  0.70] at ( 49.20,172.57) {Nemotron 8B};

\node[text=drawColor,anchor=base east,inner sep=0pt, outer sep=0pt, scale=  0.70] at ( 49.20,188.65) {Qwen3 8B};

\node[text=drawColor,anchor=base east,inner sep=0pt, outer sep=0pt, scale=  0.70] at ( 49.20,204.73) {Falcon3 7B};

\node[text=drawColor,anchor=base east,inner sep=0pt, outer sep=0pt, scale=  0.70] at ( 49.20,220.81) {Mistral 7B};

\node[text=drawColor,anchor=base east,inner sep=0pt, outer sep=0pt, scale=  0.70] at ( 49.20,236.89) {Granite 8B};
\end{scope}
\begin{scope}
\path[clip] (  0.00,  0.00) rectangle (325.21,252.94);
\definecolor{drawColor}{gray}{0.30}

\node[text=drawColor,anchor=base,inner sep=0pt, outer sep=0pt, scale=  0.64] at ( 65.00, 44.76) {0.0};

\node[text=drawColor,anchor=base,inner sep=0pt, outer sep=0pt, scale=  0.64] at (144.98, 44.76) {0.2};

\node[text=drawColor,anchor=base,inner sep=0pt, outer sep=0pt, scale=  0.64] at (224.96, 44.76) {0.4};

\node[text=drawColor,anchor=base,inner sep=0pt, outer sep=0pt, scale=  0.64] at (304.94, 44.76) {0.6};
\end{scope}
\begin{scope}
\path[clip] (  0.00,  0.00) rectangle (325.21,252.94);
\definecolor{drawColor}{RGB}{0,0,0}

\node[text=drawColor,anchor=base,inner sep=0pt, outer sep=0pt, scale=  0.80] at (187.01, 36.01) {Representation shift};
\end{scope}
\begin{scope}
\path[clip] (  0.00,  0.00) rectangle (325.21,252.94);
\definecolor{fillColor}{RGB}{231,76,60}

\path[fill=fillColor] (123.32,  8.52) rectangle (136.74, 21.94);
\end{scope}
\begin{scope}
\path[clip] (  0.00,  0.00) rectangle (325.21,252.94);
\definecolor{fillColor}{RGB}{189,195,199}

\path[fill=fillColor] (188.44,  8.52) rectangle (201.86, 21.94);
\end{scope}
\begin{scope}
\path[clip] (  0.00,  0.00) rectangle (325.21,252.94);
\definecolor{drawColor}{RGB}{0,0,0}

\node[text=drawColor,anchor=base west,inner sep=0pt, outer sep=0pt, scale=  0.70] at (141.26, 12.82) {Demo content};
\end{scope}
\begin{scope}
\path[clip] (  0.00,  0.00) rectangle (325.21,252.94);
\definecolor{drawColor}{RGB}{0,0,0}

\node[text=drawColor,anchor=base west,inner sep=0pt, outer sep=0pt, scale=  0.70] at (206.38, 12.82) {Prompt length};
\end{scope}
\end{tikzpicture}
\caption{Shift decomposition for 12~models. Grey = shift from prompt length (random-text control). Red = shift from demonstration content (content delta). Llama~3.3~70B's high total shift (0.319) is 79\% length artifact; Granite~8B's (0.610) is 86\% content. Without the random-text control, the two are indistinguishable.}
\label{fig:decomposition}
\end{figure}

If the distortion hypothesis is correct, models with larger representation shift under few-shot prompting should degrade more. Figure~\ref{fig:decomposition} reveals why this prediction fails: few-shot prompts are 5--8$\times$ longer than zero-shot prompts, and the length difference alone drives a large fraction of the observed shift.

Table~\ref{tab:main} quantifies the decomposition across all models. The proportion of content-driven shift ranges from 14\% (Gemma~3~12B) to 86\% (Granite~8B). Critically, the two degrading models -- Llama~3.3~70B and Qwen3~14B -- have the lowest content deltas among Transformers with valid behavioral data (0.067 and 0.168, respectively). Their representations change substantially under few-shot prompting, but almost none of that change comes from demonstration content.

\begin{table*}[t]
\centering
\small
\caption{Behavioral results and internal metrics. Models sorted by content delta ($\Delta_{\text{cnt}}$). Shift = total representation shift (Eq.~\ref{eq:shift}). Sh$_{\text{rnd}}$ = shift from random text (Eq.~\ref{eq:shift_rand}). $\Delta_{\text{cnt}}$ = content delta (Eq.~\ref{eq:content_delta}). Bold = degradation. Internal metrics measured on SIB-200.}
\label{tab:main}
\begin{tabular}{llR{0.9cm}R{0.9cm}R{0.7cm}R{0.8cm}R{0.8cm}R{0.8cm}}
\toprule
& & \textbf{$\Delta_{\text{FS}}$} & \textbf{$\Delta_{\text{FS}}$} & & & & \\
\textbf{Model} & \textbf{Arch} & \textbf{news} & \textbf{legal} & \textbf{DAR} & \textbf{Shift} & \textbf{Sh$_{\text{rnd}}$} & \textbf{$\Delta_{\text{cnt}}$} \\
\midrule
Granite 8B       & Trans & $+$23.5 & $+$3.1  & .442 & .610 & .087 & .523 \\
Mistral 7B       & Trans & $+$47.5 & $+$5.1  & .598 & .451 & .205 & .246 \\
Falcon3 7B       & Trans & $+$29.4 & $+$9.7  & .449 & .327 & .103 & .224 \\
Qwen3 8B         & Trans & $+$37.3 & $+$9.7  & .380 & .325 & .106 & .219 \\
Nemotron 8B      & Trans & $+$26.0 & $+$1.0  & .441 & .277 & .064 & .213 \\
Nemo 12B         & Trans & $+$26.0 & $+$1.0  & .721 & .219 & .047 & .172 \\
Qwen3 14B        & Trans & $+$35.0 & $\mathbf{-0.5}$ & .403 & .201 & .033 & .168 \\
Qwen3 32B        & Trans & $+$14.7 & $+$1.5  & .319 & .157 & .017 & .139 \\
DeepSeek R1      & Trans & ---     & ---     & .508 & .295 & .179 & .116 \\
Gemma 3 12B      & Trans & $+$2.5  & $+$4.6  & .536 & .109 & .014 & .096 \\
Llama 3.3 70B    & Trans & $\mathbf{-2.5}$ & $\mathbf{-1.0}$ & .534 & .319 & .252 & .067 \\
Falcon Mamba     & SSM   & ---     & ---     & N/A  & .035 & .007 & .027 \\
\bottomrule
\end{tabular}
\end{table*}

\subsection{Content Delta Reverses the Distortion Prediction}
\label{sec:content_delta}

\begin{figure}[t]
\centering
\input{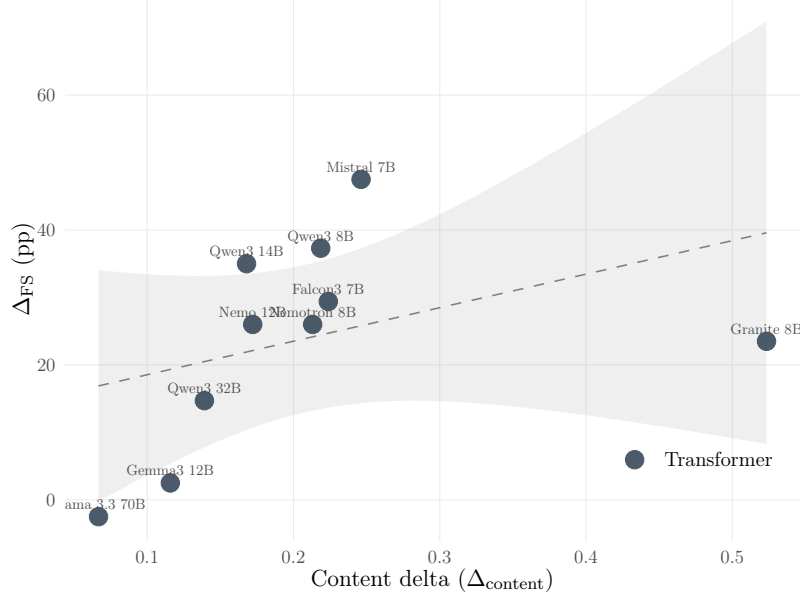}
\caption{Content delta vs.\ few-shot effect on SIB-200. Spearman $\rho = +0.648$, $p = 0.043$. The distortion hypothesis predicts a negative correlation; the data show the opposite.}
\label{fig:content_delta}
\end{figure}

Figure~\ref{fig:content_delta} plots content delta against few-shot benefit. On SIB-200, the correlation is positive and significant: $\rho = +0.648$, $p = 0.043$, $n = 10$. Models that restructure representations more from demonstration content benefit more from few-shot prompting. On ua-case-outcome, the trend is consistent but borderline ($\rho = +0.588$, $p = 0.074$, $n = 10$), likely reflecting the compressed range of legal $\Delta_{\text{FS}}$ ($-$1 to $+$10~pp vs.\ $-$2.5 to $+$47.5~pp on news).

This is the central result: \textbf{the distortion hypothesis predicts a negative correlation between representation change and few-shot benefit; the data show a positive one.} The sign reversal occurs because raw shift conflates two signals: a length artifact (which does not predict benefit; $r = +0.20$, $p = 0.59$) and a content signal (which does). Without the random-text control, the length artifact dominates, masking the true relationship.

\paragraph{Robustness.} The correlation depends partly on Granite~8B ($\Delta_{\text{cnt}} = 0.523$, the highest-delta model): excluding it weakens $\rho$ from 0.65 to 0.50. A leave-one-out analysis shows that excluding any single model changes $\rho$ by at most 0.15, with Granite having the largest influence. We report this transparently; the correlation is suggestive at $n = 10$ and would benefit from validation on additional models and tasks.

\paragraph{DAR does not predict.} The Demonstration Attention Ratio -- another intuitive candidate for explaining few-shot degradation -- shows zero correlation with $\Delta_{\text{FS}}$ ($r = +0.02$, $p = 0.96$). Nemo~12B allocates the most attention to demonstrations (DAR = 0.72) yet improves by $+$26~pp. DAR reflects how much a model attends to demonstrations, not whether that attention helps or hurts.

\subsection{Causal Confirmation}
\label{sec:causal}

To test whether demonstrations \emph{causally} harm Llama~3.3~70B -- the model with the lowest content delta -- we zero the attention weights from the final token to all demonstration tokens at every layer and head. This forces classification based solely on the input text and system prompt while preserving the positional encoding of the original few-shot sequence.

\begin{figure}[t]
\centering
\begin{tikzpicture}[x=1pt,y=1pt]
\definecolor{fillColor}{RGB}{255,255,255}
\path[use as bounding box,fill=fillColor,fill opacity=0.00] (0,0) rectangle (252.94,180.67);
\begin{scope}
\path[clip] (  0.00,  0.00) rectangle (252.94,180.67);
\definecolor{fillColor}{RGB}{255,255,255}

\path[fill=fillColor] (  0.00,  0.00) rectangle (252.94,180.68);
\end{scope}
\begin{scope}
\path[clip] ( 23.06, 22.32) rectangle (248.94,176.67);
\definecolor{drawColor}{gray}{0.92}

\path[draw=drawColor,line width= 0.4pt,line join=round] ( 23.06, 29.33) --
	(248.94, 29.33);

\path[draw=drawColor,line width= 0.4pt,line join=round] ( 23.06, 76.11) --
	(248.94, 76.11);

\path[draw=drawColor,line width= 0.4pt,line join=round] ( 23.06,122.88) --
	(248.94,122.88);

\path[draw=drawColor,line width= 0.4pt,line join=round] ( 23.06,169.66) --
	(248.94,169.66);
\definecolor{drawColor}{RGB}{44,62,80}

\path[draw=drawColor,line width= 0.5pt,dash pattern=on 4pt off 4pt ,line join=round] ( 23.06,107.26) -- (248.94,107.26);

\node[text=drawColor,anchor=base,inner sep=0pt, outer sep=0pt, scale=  0.57] at (208.87,112.78) {ZS baseline};
\end{scope}
\begin{scope}
\path[clip] (  0.00,  0.00) rectangle (252.94,180.67);
\definecolor{drawColor}{gray}{0.30}

\node[text=drawColor,anchor=base east,inner sep=0pt, outer sep=0pt, scale=  0.64] at ( 19.46, 27.13) {50};

\node[text=drawColor,anchor=base east,inner sep=0pt, outer sep=0pt, scale=  0.64] at ( 19.46, 73.90) {55};

\node[text=drawColor,anchor=base east,inner sep=0pt, outer sep=0pt, scale=  0.64] at ( 19.46,120.68) {60};

\node[text=drawColor,anchor=base east,inner sep=0pt, outer sep=0pt, scale=  0.64] at ( 19.46,167.45) {65};
\end{scope}
\begin{scope}
\path[clip] (  0.00,  0.00) rectangle (252.94,180.67);
\definecolor{drawColor}{gray}{0.30}

\node[text=drawColor,anchor=base,inner sep=0pt, outer sep=0pt, scale=  0.64] at ( 44.92, 14.31) {0};

\node[text=drawColor,anchor=base,inner sep=0pt, outer sep=0pt, scale=  0.64] at ( 81.36, 14.31) {0.1};

\node[text=drawColor,anchor=base,inner sep=0pt, outer sep=0pt, scale=  0.64] at (117.79, 14.31) {0.25};

\node[text=drawColor,anchor=base,inner sep=0pt, outer sep=0pt, scale=  0.64] at (154.22, 14.31) {0.5};

\node[text=drawColor,anchor=base,inner sep=0pt, outer sep=0pt, scale=  0.64] at (190.65, 14.31) {1};

\node[text=drawColor,anchor=base,inner sep=0pt, outer sep=0pt, scale=  0.64] at (227.09, 14.31) {2};
\end{scope}
\begin{scope}
\path[clip] (  0.00,  0.00) rectangle (252.94,180.67);
\definecolor{drawColor}{RGB}{0,0,0}

\node[text=drawColor,anchor=base,inner sep=0pt, outer sep=0pt, scale=  0.80] at (136.00,  5.56) {Demo attention scale};
\end{scope}
\begin{scope}
\path[clip] (  0.00,  0.00) rectangle (252.94,180.67);
\definecolor{drawColor}{RGB}{0,0,0}

\node[text=drawColor,rotate= 90.00,anchor=base,inner sep=0pt, outer sep=0pt, scale=  0.80] at (  9.51, 99.50) {Accuracy (\%)};
\end{scope}
\end{tikzpicture}
\caption{Attention masking in Llama~3.3~70B. Zeroing attention to demonstration tokens recovers accuracy above the zero-shot baseline, confirming that demonstration \emph{content} -- not demonstration \emph{presence} -- causes the degradation.}
\label{fig:ablation}
\end{figure}
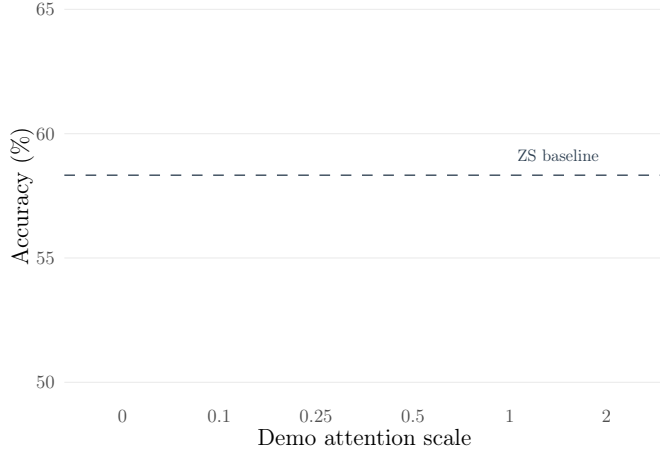

Masking recovers accuracy from 55.9\% (FS) to 59.3\% (Figure~\ref{fig:ablation}), exceeding the zero-shot baseline (58.3\%) by 1~pp. The improvement beyond zero-shot suggests that the expanded context window provides useful positional structure even when demonstration content is removed. The demonstrations actively harm this model: their content degrades performance, but their length helps.

This result is currently specific to one model (Llama~3.3~70B, 4-bit quantized). We cannot determine whether the causal harm originates from quantization artifacts or the base architecture. Extending this ablation to non-quantized models that degrade (e.g., Qwen3~14B on legal) is a priority for future work.

\subsection{Cross-Task Generalization}
\label{sec:cross_task}

Internal metrics were measured on SIB-200, yet content delta also predicts few-shot benefit on ua-case-outcome ($\rho = +0.59$, $p = 0.074$). This cross-task prediction suggests that content delta captures a \emph{model property} -- how effectively a model extracts task-relevant signal from demonstrations -- rather than a task-specific artifact.


\section{Discussion}

\subsection{Why Distortion Fails as an Explanation}

The distortion hypothesis makes a clear prediction: models whose representations shift more under few-shot prompting should perform worse. This prediction fails ($r = +0.20$) because it treats all representation change as equal. Our decomposition reveals two qualitatively different components:

\begin{enumerate}[leftmargin=*]
    \item \textbf{Length-driven shift} (captured by the random-text control): a mechanical response to longer input sequences, mediated by positional encoding and context processing. This component varies from 14\% to 79\% of total shift and carries no task-relevant information.
    \item \textbf{Content-driven shift} (content delta): restructuring caused by what demonstrations actually say. This component correlates positively with few-shot benefit, consistent with the view that ICL creates functional task vectors \citep{hendel2023icl, todd2024function} rather than noise.
\end{enumerate}

The distortion account confuses these two signals. In Llama~3.3~70B, 79\% of the shift is length-driven; the model appears ``distorted'' but is actually just responding to a longer prompt. Its content delta (0.067) is the lowest among Transformers -- it barely processes demonstration content at all, which is precisely why demonstrations do not help.

\subsection{The Random-Text Control as Standard Methodology}

Our central methodological contribution is not content delta itself but the random-text control that makes it computable. Any study that:
\begin{itemize}[leftmargin=*]
    \item measures representation shift between zero-shot and few-shot conditions,
    \item compares shift magnitude across models with different context lengths,
    \item or attributes representation change to demonstration content,
\end{itemize}
should include a random-text (or other length-matched) control. Without it, the confound between prompt length and demonstration content is unresolvable, and correlations between shift and behavior may be artifactual or reversed.

The control requires one additional forward pass per example -- negligible overhead relative to the few-shot evaluation itself. We provide implementation details and code alongside the released data.

\subsection{Architectural Notes}

Falcon Mamba~7B (SSM) shows 10$\times$ lower total shift than the Transformer mean (0.035 vs.\ 0.299). The same-provider comparison is revealing: Falcon3~7B (Transformer, shift = 0.327) vs.\ Falcon Mamba~7B (SSM, shift = 0.035). Because Falcon Mamba produced invalid outputs on both tasks, we cannot determine whether this lower shift translates to different few-shot behavior. The architecture comparison remains an observation from $n = 1$; we note it for future work with more SSM models.


\section{Limitations}

\paragraph{Sample size for correlations.} Content delta correlations are computed over $n = 10$ models. The significance on SIB-200 ($p = 0.043$) is just below the conventional threshold; on ua-case-outcome ($p = 0.074$) it is not. The finding would be strengthened by evaluating additional models and task types (generation, retrieval, reasoning).

\paragraph{Single causal ablation.} The attention-masking experiment is specific to Llama~3.3~70B (4-bit quantized). The degradation--quantization confound cannot be resolved without FP16 inference. Extending this to non-quantized degrading models is needed.

\paragraph{Random-text control validity.} Random text differs from real demonstrations in entropy, token distribution, and syntactic structure -- not only in length. The control captures prompt length but may also capture other non-content factors. Shuffled-demonstration and wrong-label controls would provide a finer-grained decomposition.

\paragraph{Two classification tasks.} Both tasks are closed-set classification with short answers. Content delta's predictive power on generation, extraction, or reasoning tasks is untested.

\paragraph{Demo truncation.} Legal demonstrations were truncated to 300~characters (from typical 5,000--30,000) to fit 7B context windows. This truncation may independently reduce few-shot benefit on legal text.

\paragraph{Greedy decoding.} Temperature~0 may systematically favor certain labels depending on tokenizer-specific token probabilities, introducing bias not present under sampling.


\section{Conclusion}

We tested the distortion hypothesis -- the assumption that representation shift from few-shot demonstrations causes degradation -- and found it wrong. Three findings:

\begin{enumerate}[leftmargin=*]
    \item \textbf{Raw representation shift does not predict few-shot benefit or harm} ($r = 0.20$, $p = 0.59$). The apparent ``distortion'' in degrading models is predominantly a prompt-length artifact: 40--79\% of observed shift comes from the longer input, not from demonstration content.

    \item \textbf{Content delta -- shift attributable to demonstration content after controlling for prompt length -- predicts benefit, not harm} ($\rho = +0.65$, $p = 0.043$). Models that restructure representations more from what demonstrations say improve more. The model that degrades most (Llama~3.3~70B) has the lowest content delta among Transformers.

    \item \textbf{The random-text control is a necessary methodological step.} Without it, prompt length inflates shift estimates by up to 5$\times$, reverses the sign of the shift--benefit correlation, and supports the wrong hypothesis. The control requires one additional forward pass and should become standard practice in representation-geometric analyses of ICL.
\end{enumerate}

\section*{Acknowledgments}

Compute was provided by AWS Activate credits: SageMaker g5.12xlarge (4$\times$A10G) for all experiments. Total compute cost: \$95. We thank the creators of SIB-200 and all model providers for open-weight access. Data and code: \url{https://huggingface.co/datasets/overthelex/attention-analysis-fewshot}.


\appendix
\section{Model Identifiers}
\label{app:models}

\begin{table}[h]
\centering
\small
\caption{HuggingFace model identifiers for reproducibility.}
\label{tab:model_ids}
\begin{tabular}{ll}
\toprule
\textbf{Model} & \textbf{HuggingFace ID} \\
\midrule
Qwen3 8B       & \texttt{Qwen/Qwen3-8B} \\
Qwen3 14B      & \texttt{Qwen/Qwen3-14B} \\
Qwen3 32B      & \texttt{Qwen/Qwen3-32B} \\
Llama 3.3 70B  & \texttt{meta-llama/Llama-3.3-70B-Instruct} \\
Mistral 7B     & \texttt{mistralai/Mistral-7B-Instruct-v0.3} \\
Nemo 12B       & \texttt{mistralai/Mistral-Nemo-Instruct-2407} \\
Nemotron 8B    & \texttt{nvidia/Llama-3.1-Nemotron-8B-Instruct} \\
Falcon3 7B     & \texttt{tiiuae/Falcon3-7B-Instruct} \\
Granite 3.1 8B & \texttt{ibm-granite/granite-3.1-8b-instruct} \\
DeepSeek R1 14B & \texttt{deepseek-ai/DeepSeek-R1-Distill-Qwen-14B} \\
Gemma 3 12B    & \texttt{google/gemma-3-12b-it} \\
Falcon Mamba 7B & \texttt{tiiuae/falcon-mamba-7b-instruct} \\
\bottomrule
\end{tabular}
\end{table}

\section{Within-Family Scaling}
\label{app:scaling}

The Qwen3 family (8B, 14B, 32B) provides a controlled within-family comparison. On SIB-200, the few-shot benefit decreases monotonically with size: $+$37.3, $+$35.0, $+$14.7~pp. On ua-case-outcome, the pattern is non-monotonic: 8B benefits most ($+$9.7~pp), 14B \emph{degrades} ($-$0.5~pp), and 32B shows a small benefit ($+$1.5~pp). Content delta also decreases with size (0.219, 0.168, 0.139), consistent with the content hypothesis: larger Qwen models extract less task-relevant signal from demonstrations, reducing the benefit. The 14B model occupies a threshold: its content delta (0.168) falls below the value needed to offset the task difficulty of legal classification.

\end{document}